\documentclass[11pt]{article}

\usepackage[final]{acl}

\usepackage{times}
\usepackage{latexsym}
\usepackage{booktabs}
\usepackage{multirow}
\usepackage[T1]{fontenc}
\usepackage{amsfonts}
\usepackage{amsthm}
\usepackage{enumitem}
\usepackage{makecell}
\usepackage{subcaption}
\usepackage{tabularx}
\usepackage{amsmath}
\usepackage{colortbl}
\usepackage[table]{xcolor} 
\usepackage{booktabs}
\usepackage{tabularx}
\usepackage{array}
\usepackage{siunitx}
\usepackage{xcolor}
\newcolumntype{C}{>{\centering\arraybackslash}X}
\usepackage{subcaption}
\usepackage[most]{tcolorbox}
\usepackage{xcolor}
\usepackage{cuted}  
\newtcolorbox{redexampleFloatingTitle}[1]{%
  enhanced,breakable,
  colframe=red!80!black,
  colbacktitle=red!80!black,
  coltitle=white,
  colback=red!2,,
  fonttitle=\bfseries\large,
  fontupper=\small,
  boxrule=1pt,
  arc=6pt,
  toptitle=2pt,bottomtitle=2pt,
  top=10pt,bottom=10pt,left=8pt,right=8pt,
  title={#1}%
}

\newcolumntype{Y}{S[table-format=2.1]}
\definecolor{Gray}{gray}{0.9} 
\definecolor{OursBlue}{HTML}{E6F7FF} 
\theoremstyle{plain}
\newtheorem{theorem}{Theorem}

\theoremstyle{definition}

\newcommand{\Solver}{\texttt{Solver}}
\newcommand{\Questioner}{\texttt{Questioner}}

\usepackage[utf8]{inputenc}

\usepackage{microtype}

\usepackage{inconsolata}

\usepackage{graphicx}

\usepackage{amsmath}
\usepackage{graphicx}
\usepackage{multirow}
\usepackage{xspace}

\newcommand{\method}{\textsc{DARC}\xspace}

\title{DARC: Decoupled Asymmetric Reasoning Curriculum for LLM Evolution}

\author{
    \textbf{Shengda Fan$^{1}$}\thanks{\; The first two authors contributed equally.}, \textbf{Xuyan Ye$^{1*}$}, \textbf{Yankai Lin$^{1}$}\thanks{\; Corresponding author.} \\
  $^1$ Gaoling School of Artificial Intelligence, Renmin University of China \\
  \texttt{\{fanshengda,  yexvyan0923,  yankailin\}@ruc.edu.cn}
}

\begin{document}
\maketitle

\maketitle
\begin{abstract}
Self-play with large language models has emerged as a promising paradigm for achieving self-improving artificial intelligence.
However, existing self-play frameworks often suffer from optimization instability, due to (i) non-stationary objectives induced by solver-dependent reward feedback for the \Questioner{}, and (ii) bootstrapping errors from self-generated pseudo-labels used to supervise the \Solver{}.
To mitigate these challenges, we introduce \textbf{\method{}} (\textbf{D}ecoupled \textbf{A}symmetric \textbf{R}easoning \textbf{C}urriculum), a two-stage framework that stabilizes the self-evolution process.
First, we train the \Questioner{} to synthesize difficulty-calibrated questions, conditioned on explicit difficulty levels and external corpora.
Second, we train the \Solver{} with an asymmetric self-distillation mechanism, where a document-augmented teacher generates high-quality pseudo-labels to supervise the student \Solver{} that lacks document access. 
Empirical results demonstrate that \method{} is model-agnostic, yielding an average improvement of \textbf{10.9} points across nine reasoning benchmarks and three backbone models. Moreover, \method{} consistently outperforms all baselines and approaches the performance of fully supervised models without relying on human annotations.
The code is available at \url{https://github.com/RUCBM/DARC}.
\end{abstract}

\section{Introduction}

Self-improving artificial intelligence, which enables models to autonomously refine their capabilities without human intervention, is widely viewed as an important step toward more general and potentially superhuman intelligence~\cite{schmidhuber2007godel}.
While large language models (LLMs) have demonstrated remarkable progress in complex reasoning tasks~\cite{luo2023wizardmath, liu2025logical}, their success largely relies on extensive human supervision.
However, as the availability of high-quality human-annotated data approaches its limit, this scarcity becomes a fundamental bottleneck for further scaling. Consequently, developing reliable and efficient self-evolution mechanisms that operate independently of human data has emerged as a pivotal research frontier.

\begin{figure}[t]
    \centering
    \includegraphics[width=1.0 \columnwidth]{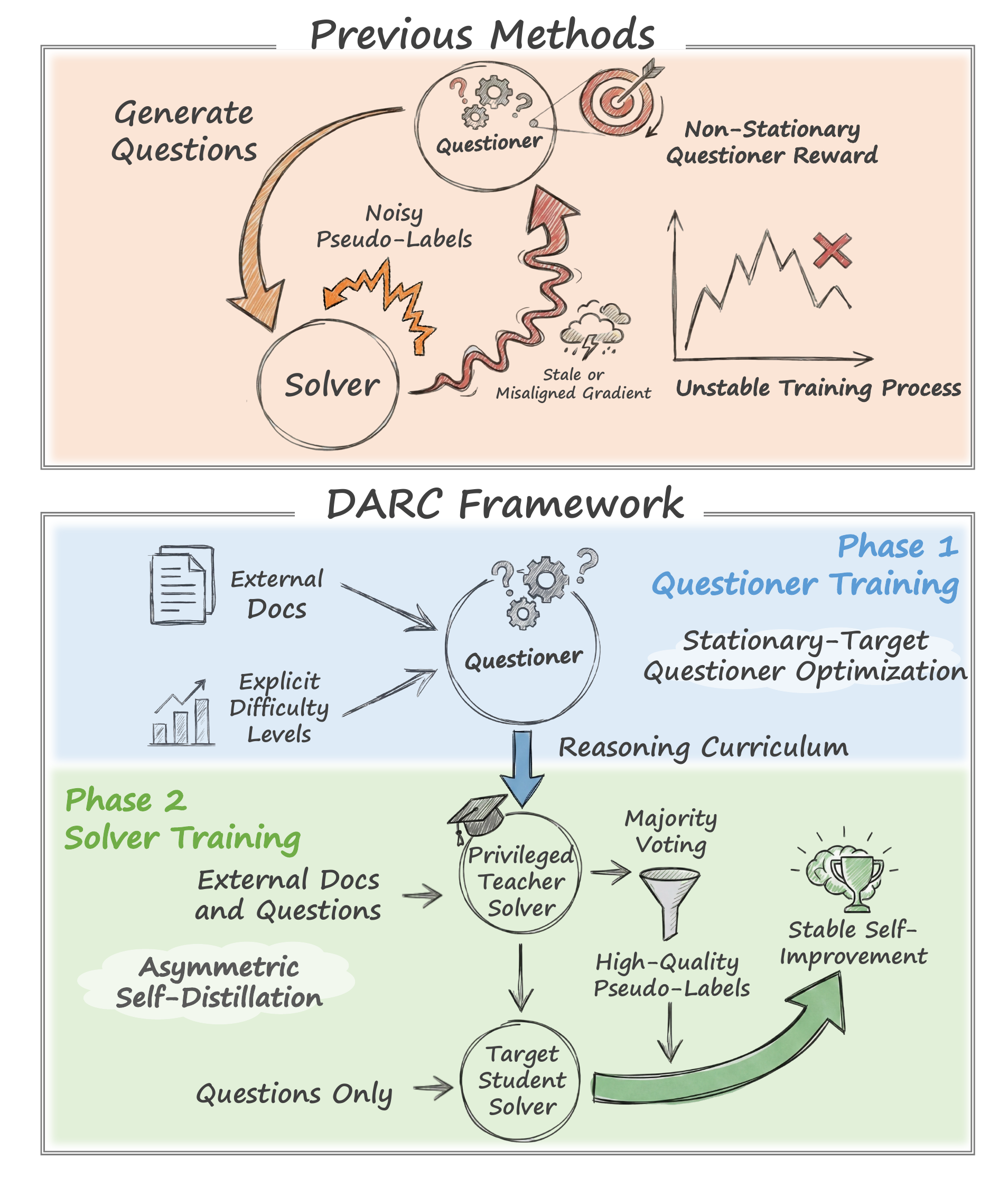}
    \caption{Comparison between \method{} and previous coupled self-play methods.}
    \label{fig::teaser}
\end{figure}

A widely adopted approach for LLM self-evolution is self-play, which establishes a \textbf{co-evolutionary loop} between a \Questioner{} that proposes tasks and a \Solver{} that attempts to solve them~\cite{silver2017mastering,sukhbaatar2017intrinsic}. In this paradigm, the \Questioner{} optimizes a difficulty-calibration reward based on the \Solver{}'s current performance, while the \Solver{} is updated on pseudo-labeled data constructed from the tasks generated by the \Questioner{}.
Despite its appeal, such tightly coupled systems are prone to premature plateaus or even performance collapse~\citep{huang2025r}. 
We argue that this instability stems from two inherent issues in the coupled design: \textbf{non-stationary optimization targets} and \textbf{self-confirmation bias}. 
As shown in the upper part of Figure \ref{fig::teaser}, the \Questioner{} is continually optimized against a moving target determined by the evolving \Solver{}. Its learned difficulty signal can become stale or even misaligned after each \Solver{} update, inducing oscillatory shifts in task difficulty and unstable optimization dynamics. In parallel, the \Solver{} is trained on self-generated pseudo-labels that are inevitably noisy, so errors can be propagated and amplified across iterations.
%

To address the coupled instability in self-play, we propose \textbf{D}ecoupled \textbf{A}symmetric \textbf{R}easoning \textbf{C}urriculum (\method{}), which restructures self-evolution into two decoupled and sequential stages: \Questioner{} training and \Solver{} training, as illustrated in the lower part of Figure~\ref{fig::teaser}.
In Stage~1, instead of chasing a moving \Solver{} boundary, we train the \Questioner{} with \emph{objective} supervision from explicit difficulty levels and external documents, enabling it to generate corpus-grounded questions that are calibrated to a specified difficulty.
This decoupling removes the dependence of question generation on the \Solver{}'s real-time performance, thereby mitigating non-stationary optimization targets.
In Stage~2, we replace self-training on noisy pseudo-labels with an \emph{asymmetric self-distillation} scheme.
Concretely, we introduce a privileged teacher \Solver{} that has access to the source document and performs majority voting to construct pseudo-labels, and distill its outputs into a student \Solver{} that receives only the question as input.
This asymmetric design (document-augmented teacher vs.\ question-only student) reduces label noise and alleviates self-confirmation bias, yielding a more stable and reliable learning signal for \Solver{} improvement.


Our experiments demonstrate that \method{} is model-agnostic, consistently improving the reasoning abilities of both Qwen-based~\cite{yang2025qwen3} and LLaMA-based~\cite{wang2025octothinker} backbones. Notably, applying \method{} improves the average accuracy by \textbf{10.9} points over the base models, outperforming label-free self-evolving methods R-Zero~\cite{huang2025r} and Absolute-Zero~\cite{zhao2025absolute}, corpus-grounded self-play method SPICE \cite{liu2025spice} and weakly supervised self-play methods R-Few \cite{yu2025guided}.
Notably, \method{} approaches the performance of General-Reasoner~\cite{ma2025general}, which is trained on the full 232K WebInstruct dataset~\cite{yue2024mammoth2}, despite using no human annotations.
Beyond empirical gains, we further demonstrate several key properties of \method{}:
(i) it does not merely memorize the training corpus;
(ii) the difficulty rankings are solver-independent; and
(iii) the decoupled \Questioner{} learns a curriculum improving the performance of heterogeneous \Solver{} backbones.

\section{Related Work}

\paragraph{Reinforcement Learning for LLM Reasoning.}
Reinforcement learning (RL) is pivotal for advancing LLM reasoning, but a key bottleneck lies in how to obtain reliable supervision signals. 
A primary paradigm employs \emph{external verifiers}~\citep{guo2025deepseek} to provide rule-based rewards, but this approach is largely restricted to deterministic fields like mathematics and code. Alternatively, \emph{self-supervision} methods derive signals via maximizing confidence~\citep{prabhudesai2025maximizing} or getting pseudo-labels by majority voting~\citep{zuo2025ttrl}, yet they remain dependent on curated question sets. 
In contrast, \method{} transcends these limitations by operating  on raw corpora, generating both tasks and supervision signals in a self-evolving manner.


\paragraph{Self-Play for LLM Self-Evolution.}
Self-play is a widely adopted paradigm for enabling self-evolution in LLMs~\cite{chen2024self}.
One line of research focuses on code-centric self-play, where execution results provide grounded supervision and enable stable co-evolution between a coder and a verifier \citep{zhao2025absolute, wang2025cure, lin2025learning}. 
A parallel line extends self-play to general reasoning via a questioner--solver framework, in which a \Questioner{} proposes tasks and a \Solver{} learns from self-generated supervision \citep{huang2025r, liu2025spice, chen2025self}.
Unlike code-centric settings that can rely on executable feedback for supervision, general-reasoning self-play often suffers from non-stationary rewards and unstable optimization.
In contrast, our approach stabilizes self-evolving optimization by decoupling training and introducing asymmetric self-distillation.


\paragraph{Data Synthesis for LLM Training.} Synthetic data generation is pivotal for scaling reasoning supervision. Bootstrapping frameworks, including STaR \citep{zelikman2022star}, ReST \citep{gulcehre2023reinforced}, and WizardLM \cite{xu2024wizardlm}, expand datasets by iteratively evolving instructions from seed exemplars. Meanwhile, corpus-driven pipelines like WebInstruct \citep{yue2024mammoth2} and General-Reasoner \citep{ma2025general} synthesize question-answer pairs by leveraging web mining or model priors to broaden domain coverage. While these approaches effectively scale data quantity or diversity, they often rely on superior external models (e.g., GPT-4o, Gemini) to guarantee data quality. 
In contrast, \method{} dispenses with external teacher models by deriving both task difficulty and supervision signals internally, enabling stable self-evolution directly from unlabeled corpora.



\section{Methodology}
\begin{figure*}[htbp]
  \centering
  \includegraphics[width=\textwidth]{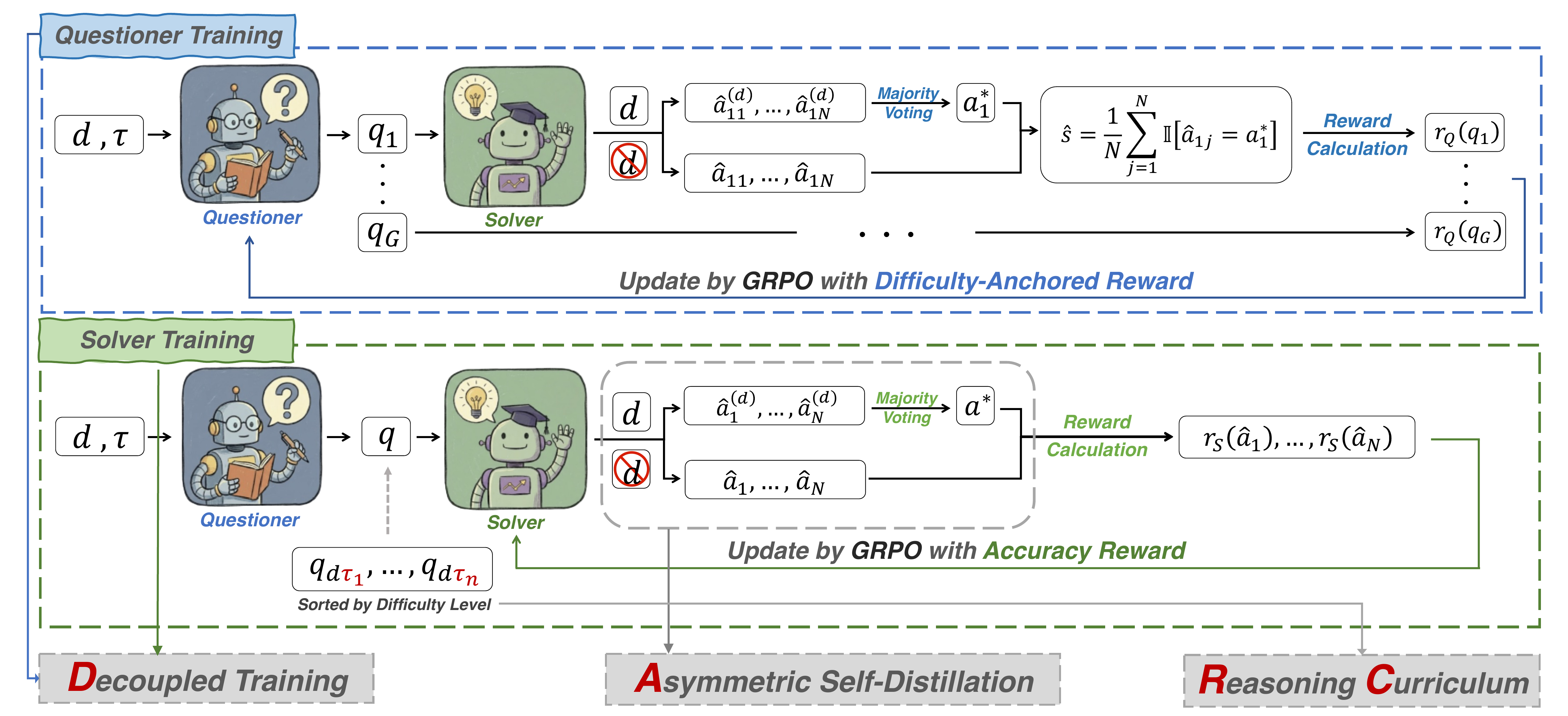}
  \caption{
 Illustration of the two-stage \method{} framework.
In the first stage (the upper half), the \Questioner{} learns to generate questions matching specified difficulty $\tau$  with a difficulty-anchored reward. In the second stage (the lower half), the \Solver{} is trained on an offline curriculum with an answer correctness reward.
  }
  \label{fig:pipeline}
\end{figure*}
In this section, we present the \method{} framework, which decomposes LLM self-evolution into two sequential stages.
In the first stage (Section~\ref{sec:questioner}), 
the \Questioner{} is trained for controllable question generation based on explicit difficulty levels and an external corpus.
In the second stage (Section~\ref{sec:solver}), the trained \Questioner{} is used to construct an offline reasoning curriculum for training the \Solver{} via asymmetric self-distillation.
Finally, in Section~\ref{sec:theory}, we provide a theoretical analysis that explains how the proposed decoupling and training strategy mitigates optimization instability in self-evolving systems.
Figure~\ref{fig:pipeline} visually outlines the \method{} framework.


\subsection{\Questioner{}: Difficulty-Aware Generation}
\label{sec:questioner}
Given a document $d$ sampled from the corpus $\mathcal{D}$ and a target difficulty scalar $\tau \in [0,1]$, the \Questioner{} $\mathcal{Q}_\theta$ defines a conditional generation policy
$q \sim \mathcal{Q}_\theta(\cdot \mid d, \tau)$,
which aims to produce questions grounded in $d$ and calibrated to the specified difficulty level.

To train the \Questioner{}, we adopt Group Relative Policy Optimization (GRPO)~\cite{shao2024deepseekmath} to optimize the policy using a reward function that jointly enforces document grounding and difficulty alignment.
Concretely, for each generated question $q$, the reward is computed through a two-stage evaluation procedure.
First, we employ an LLM-as-a-Judge to verify whether the question $q$ is grounded in the source document $d$.
Questions that fail this grounding check are assigned a negative reward to discourage hallucinated or irrelevant generations.
Second, for grounded questions, we estimate their empirical difficulty using a fixed \Solver{} $\mathcal{S}(\cdot \mid q)$.
Specifically, we sample $N$ candidate answers $\{\hat{a}_1, \dots, \hat{a}_N\}$ 
from a fixed base model serving as the \Solver{}, and compute the empirical success rate
\begin{equation} \label{empirical_success_rate}
    \hat{s} = \frac{1}{N}\sum_{j=1}^N \mathbb{I}[\hat{a}_j = a^*],
\end{equation}
where $a^*$ denotes the pseudo-label obtained via majority voting from a document-augmented \Solver{} (see Section~\ref{sec:solver}). We denote this empirical success rate as the difficulty estimator $D(q) = \hat{s}$.

The final reward for question $q$ is then defined as
\begin{equation}
    r_Q(q) =
    \begin{cases}
    1 - |D(q) - \tau|, & \text{if } q \text{ is grounded in } d, \\
    -1, & \text{otherwise}.
    \end{cases}
    \label{eq:reward}
\end{equation}

This reward formulation penalizes ungrounded questions while encouraging the generated questions to match the target difficulty $\tau$.

\subsection{\Solver{}: Offline Curriculum Learning via Asymmetric Self-Distillation}
\label{sec:solver}
After training the \Questioner{}, we freeze its parameters and construct an offline question set
\begin{equation}
\mathcal{U}=\{(d_i,\tau_i,q_i)\}_{i=1}^{M}, 
\quad q_i\sim\mathcal{Q}_{\theta}(\cdot\mid d_i,\tau_i).
\end{equation}
We adopt curriculum learning~\cite{bengio2009curriculum} by ordering questions from easy to hard according to the specified difficulty level $\tau_i$, and train the \Solver{} progressively along this curriculum.

To obtain supervision without external annotations, we employ asymmetric self-distillation.
For each $(d,q)\in \mathcal{U}$, we introduce a privileged teacher \Solver{} with access to the source document, which generates multiple candidate answers:
\begin{equation}
\hat{a}^{(d)}_1,\ldots,\hat{a}^{(d)}_N \sim \mathcal{S}_{\phi}(\cdot\mid d,q).
\end{equation}
We obtain a pseudo-label $a^*$ by majority voting over $\{\hat{a}^{(d)}_i\}_{i=1}^{N}$.
We discard samples with vote agreement below $\gamma$ to reduce label noise.
The remaining samples are used to train a student \Solver{} $\mathcal{S}_{\phi}(\cdot\mid q)$, which shares parameters with the privileged teacher but doesn't have access to $d$.
This asymmetry discourages trivial copying from the document and reduces confirmation bias, forcing the student to learn to solve problems from questions alone.
We optimize the student \Solver{} with a correctness reward:
\begin{equation}
r_S(a)=\mathbb{I}[a=a^*],\quad a \sim \mathcal{S}_{\phi}(\cdot\mid q).
\end{equation}

Together, this offline curriculum learning scheme with asymmetric self-distillation provides a stable and scalable training signal, enabling effective optimization without external supervision.

\subsection{Theoretical Analysis}
\label{sec:theory}
We provide a self-contained theoretical analysis to elucidate why \emph{coupled self-play} is inherently unstable from an optimization perspective, and how the proposed decoupling strategy alleviates the issue.

\subsubsection{Toy Model of Coupled Self-Play}
We consider a simplified one-dimensional abstraction that captures the essential dynamics of difficulty-controlled self-play.

\paragraph{Setup.}
For theoretical tractability, each question is characterized by a scalar difficulty $\tau \in \mathbb{R}$ (relaxed from $[0,1]$), where larger values indicate harder questions.
The \Solver{} is parameterized by a scalar ability parameter $\phi_t \in \mathbb{R}$ at iteration $t$.
The probability that the \Solver{} successfully answers a question of difficulty $\tau$ is defined as
\begin{equation}
v_{\phi_t}(\tau) := \sigma(\phi_t - \tau),
\end{equation}
where $\sigma(z) = (1 + e^{-z})^{-1}$ is the logistic function.

The \Questioner{} is parameterized by $\theta$ and induces questions of a distribution over difficulties,
\begin{equation}
\tau \sim \pi_\theta(\cdot).
\label{eq:toy_difficulty_policy}
\end{equation}
Following common practice in self-play systems~\citep{huang2025r}, we assume the \Questioner{} is trained to generate boundary questions whose empirical success rate is close to $0.5$.
This is formalized via the shaped objective
\begin{equation}
J_t(\theta) := \mathbb{E}_{\tau \sim \pi_\theta}\Big[\psi\!\left(v_{\phi_t}(\tau)\right)\Big],
\
\psi(u) := -\lvert u - \tfrac{1}{2} \rvert.
\label{eq:toy_objective}
\end{equation}
By construction, $J_t(\theta)$ is maximized when $\pi_\theta$ concentrates its mass near $\tau=\phi_t$.

\subsubsection{Structural Instability in Coupled Self-Play}

We now show that in the above minimal setting, \Solver{} updates can cause the \Questioner{}’s reward ascent direction to become stale, such that an update that is optimal for the current objective provably harms the next objective.

\begin{theorem}[Gradient direction reversal under coupling]
\label{thm:direction_flip}
Consider the toy model above.
Assume the \Solver{} updates according to
\begin{equation}
\phi_{t+1} = \phi_t + \eta,
\qquad \eta \neq 0.
\end{equation}
Let the \Questioner{} perform a single gradient ascent step
\begin{equation}
\theta_{t+1} = \theta_t + \alpha g_t,
\qquad g_t \in \partial J_t(\theta_t),
\end{equation}
where $\partial J_t(\theta_t)$ denotes a (sub)gradient and $\alpha > 0$ is sufficiently small.

Then, for any $\delta$ satisfying
\begin{equation}
\delta \in
\begin{cases}
(0,\eta), & \text{if } \eta > 0, \\
(\eta,0), & \text{if } \eta < 0.
\end{cases}
\end{equation}
if $\pi_{\theta_t}$ concentrates around $\tau=\phi_t+\delta$ (e.g., $\Pr_{\tau\sim\pi_{\theta_t}}[|\tau-(\phi_t+\delta)|\le \epsilon]\approx 1$ for sufficiently small $\epsilon$), there exists a constant $c(\eta, \delta) > 0$ such that
\begin{equation}
J_{t+1}(\theta_{t+1}) - J_{t+1}(\theta_t)
~\le~
- c(\eta, \delta)\, \alpha
~<~ 0.
\end{equation}
That is, a \Questioner{} update that ascends $J_t$ necessarily \emph{decreases} the next-round objective $J_{t+1}$.
\end{theorem}

\paragraph{Proof sketch.}
Under the stated concentration condition, the sign of the local ascent direction is governed by the typical difficulty $\tau\approx \phi_t+\delta$.
If $\delta>0$, then $v_{\phi_t}(\tau)=\sigma(-\delta)<1/2$ and the ascent direction pushes $\pi_\theta$ toward smaller difficulties (easier questions).
If $\delta<0$, then $v_{\phi_t}(\tau)=\sigma(-\delta)>1/2$ and the ascent direction pushes $\pi_\theta$ toward larger difficulties and harder questions.
After the \Solver{} update, we have $\phi_{t+1}-\tau = \eta-\delta$, which changes sign whenever $\delta$ and $\eta$ have the same sign and $|\delta|<|\eta|$, implying
\begin{equation}
v_{\phi_{t+1}}(\tau) = \sigma(\eta-\delta),
\end{equation}
which lies on the opposite side of $\tfrac{1}{2}$ compared to
$v_{\phi_t}(\tau)$.
Therefore, the previous ascent direction becomes a descent direction for $J_{t+1}$, i.e.,
$\langle \nabla J_{t+1}(\theta_t), g_t\rangle < 0$.
Applying a first-order Taylor expansion around $\theta_t$ gives
\begin{equation}
J_{t+1}(\theta_{t+1})-J_{t+1}(\theta_t)
=
\alpha\langle \nabla J_{t+1}(\theta_t), g_t\rangle + O(\alpha^2).
\end{equation}
Since the directional derivative is strictly negative under the sign flip condition, choosing $\alpha>0$ sufficiently small ensures the $O(\alpha^2)$ term is dominated, yielding
\begin{equation}
J_{t+1}(\theta_{t+1}) - J_{t+1}(\theta_t)
~\le~
- c(\eta, \delta)\, \alpha
~<~ 0,
\end{equation}
where $c(\eta,\delta)>0$.
\hfill$\square$

\paragraph{Interpretation.}
Theorem~\ref{thm:direction_flip} formalizes a structural instability in coupled self-play:
even when the \Questioner{} follows an ascent direction of its current objective, \Solver{} learning can immediately invalidate this direction.
This phenomenon is not caused by stochastic noise or large learning rates, but by objective drift induced by \Solver{} updates.
As a result, proposer gradients become intrinsically stale, making stable optimization difficult without severely restricting \Solver{} learning.

\begin{figure}[t]
    \centering
    \includegraphics[width=0.9\columnwidth]{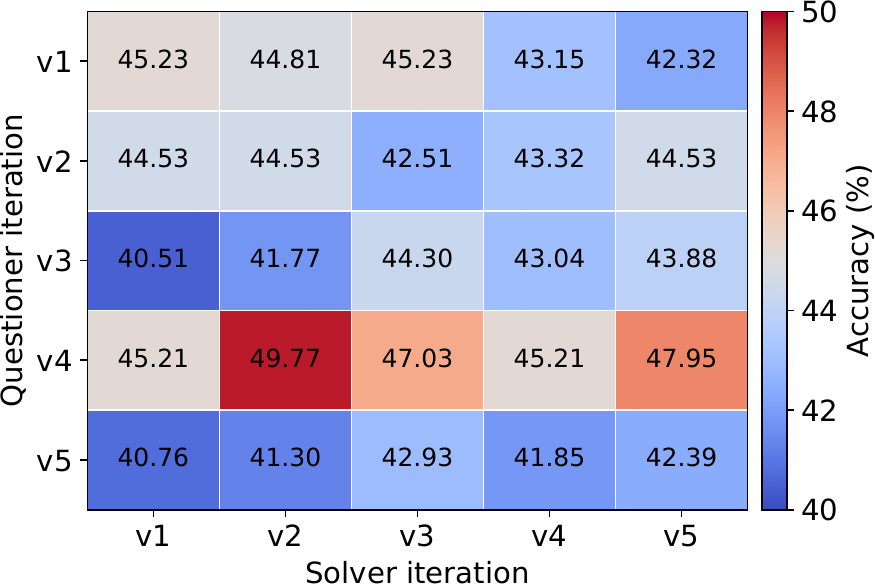}
    \caption{Cross-Iteration Accuracy Heatmap in Coupled Self-Play. Golden label is annotated by  DeepSeek-3.2~\cite{liu2025deepseek} and only for analytical purposes.}
    \label{fig::rzer_heatmap}
\end{figure}

\paragraph{Empirical Analysis.}
To probe the non-stationarity predicted by Theorem~\ref{thm:direction_flip}, we reproduce R-Zero~\cite{huang2025r}, a coupled self-play loop, that alternates (i) training a \Questioner{} $Q_i$ against the frozen \Solver{} $S_{i-1}$ and (ii) training a new \Solver{} $S_i$ on questions sampled from $Q_i$.
Figure~\ref{fig::rzer_heatmap} reports cross-iteration accuracies between the above iterations.
If the coupled loop were stable, we would expect a coherent structure: for a fixed \Solver{} (each column), later \Questioner{}s should become systematically harder (accuracy decreasing with $i$), and for a fixed \Questioner{} (each row), later \Solver{}s should improve (accuracy increasing with $t$).
The heatmap shows pronounced \textbf{non-monotonicity along both axes}, indicating instability in the coupled self-play loop.
Diagonal accuracies remain below the $0.5$ boundary (mean $\approx 0.443$) with no convergence trend, while row-wise means oscillate: after $Q_3$ becomes harder, $Q_4$ turns uniformly easier for all \Solver{}s (showing a bright horizontal band), followed by a harder $Q_5$.
This behavior aligns with solver-induced objective drift: since the \Questioner{} objective $J_t(\theta)$ is defined relative to a moving \Solver{} boundary $v_{\phi_t}(\tau)$, \Solver{} updates can invalidate or reverse previous ascent directions, producing non-monotonic and solver-specific difficulty shifts, rather than a globally ordered and progressively aligned curriculum.

\subsubsection{Why Decoupling Resolves the Instability}
In \method{}, the \Questioner{} is trained using a difficulty-matching objective
\begin{equation}
\widetilde{J}(\theta)
=
\mathbb{E}_{(d,\tau)}\,
\mathbb{E}_{q \sim \pi_\theta(\cdot \mid d, \tau)}
\big[\, \rho(D(q), \tau) \,\big],
\end{equation}
where $D(q)$ is a fixed difficulty estimator and $\rho$ encourages $D(q)$ to match the target difficulty $\tau$.
Crucially, $\widetilde{J}$ doesn't depend on the time-varying \Solver{}.
Therefore, gradient ascent steps on $\widetilde{J}$ do not suffer from the direction reversal phenomenon described in Theorem~\ref{thm:direction_flip}.
And the \Questioner{} can be trained stably using standard policy optimization methods, and difficulty progression is controlled explicitly via curriculum scheduling rather than implicitly through unstable self-play dynamics.

\begin{table*}[t]
\centering
\small
\vspace{-0.1in}
\setlength{\tabcolsep}{1.4mm}
\resizebox{\textwidth}{!}{%
\begin{tabular}{lcccccc|ccccc|c}
\toprule
\multirow{3}{*}{\textbf{Models}} 
& \multicolumn{6}{c|}{\textbf{Mathematical Reasoning}} 
& \multicolumn{5}{c|}{\textbf{General Reasoning}} 
& \multicolumn{1}{c}{\textbf{}} \\
\cmidrule{2-13}
& AMC & Minerva & MATH & GSM8K & Olympiad & \makecell[c]{\textbf{Math}\\[-1pt]\textbf{Avg.}}
&  \makecell[c]{MMLU\\[-1pt]Pro} & \makecell[c]{Super\\[-1pt]GPQA} & \makecell[c]{GPQA\\[-1pt]Diamond} & BBEH
& \makecell[c]{\textbf{General}\\[-1pt]\textbf{Avg.}} & \textbf{Avg.} \\
\midrule
\multicolumn{7}{@{}l|}{\textit{Qwen3-4B-Base}} &&&&&& \\
\rowcolor{gray!10} 
General-Reasoner 
& 60.0 & 57.7 & 80.6 & 92.2 & 47.7 & 67.6
& 62.8 & 32.5 & 42.9 & 12.2 & 37.6
& 54.3\\
Base Model 
& 47.5 & 42.3 & 68.2 & 72.6 & 34.8 & 53.1
& 51.6 & 25.4 & 26.3 & 8.1 & 27.9 
& 41.9 \\
\quad + R-Zero 
& 48.2 & 51.2 & 74.8 & 90.6 & 40.6 & 61.1
& 54.2 & 27.8 & 36.4 & 10.4 & 32.2
& 48.2 \\
\quad + Absolute Zero
& 50.0 & 41.9 & 76.2 & 89.3 & 41.5 & 59.8
& 52.6 & 27.1 & 35.3 & 8.3 & 30.8
& 46.9 \\
\quad + SPICE
& 50.9 & \underline{55.5} & \textbf{77.9} & \underline{91.9} & 41.9 & \underline{63.6}
& \underline{56.5} & 28.3 & \underline{37.9} & \textbf{11.3} & \underline{33.5}
& \underline{50.2} \\

\quad + R-Few (1\%)
& \underline{52.7} & 52.1 & \underline{77.8} & \textbf{92.3} & \underline{42.4} & 63.5
& 55.9 & \textbf{29.4} & 35.4 & \underline{11.2} & 33.0
& 49.9 \\

\quad + {DARC (Ours)}
& \textbf{60.3} &  \textbf{57.7} & 77.6 & \underline{91.9} & \textbf{45.8} & \textbf{66.7}
& \textbf{56.9} & \underline{29.2} & \textbf{38.9} & \underline{11.2} & \textbf{34.1}
& \textbf{52.2} \\

\midrule
\multicolumn{7}{@{}l|}{\textit{Qwen3-8B-Base}} &&&&&& \\
\rowcolor{gray!10} 
General-Reasoner  
& 64.8 & 62.6 & 83.4 & 92.7 & 46.3 & 70.0
& 65.1 & 35.3 & 42.9 & 10.8 & 38.5
& 56.0 \\
Base Model 
& 61.5 & 49.3 & 74.4 & 90.9 & 40.4 & 63.3
& 58.0 & 30.4 & 33.3 & 10.5 & 33.1
& 49.9 \\
\quad + R-Zero 
& 62.8 & 58.8 & 80.6 & 92.4 & 43.4 & 67.6
& 61.6 & 31.8 & \underline{40.5} & 11.3 & 36.3
& 53.7 \\
\quad + Absolute Zero 
& 62.5 & 52.9 & 76.6 & 92.0 & 47.8 & 66.4
& \underline{62.5} &  \textbf{33.5} & 36.8 & 10.8 & 35.9
& 52.8 \\

\quad + SPICE
& 60.9 & 55.2 & 81.4 & 93.8 & \underline{48.0} & 67.9
& 61.0 & 32.4 & 40.4 & \textbf{12.1} & 36.5
& 53.9 \\

\quad + R-Few (1\%) 
& \textbf{69.3} & \underline{59.6} & \underline{81.6} & \textbf{94.0} & 44.0 & \underline{69.7}
&  \textbf{62.8} & 32.7 & 40.4 &  \underline{11.8} & \underline{36.9}
& \underline{55.1} \\
\quad + {DARC (Ours)}
& \underline{68.9} & \textbf{61.4} & \textbf{83.0} & \textbf{94.0} & \textbf{48.4}  & \textbf{71.1}
&  62.3 & \underline{32.8} & \textbf{44.4} &  \underline{11.8} & \textbf{37.8}
& \textbf{56.3} \\



\midrule

\multicolumn{7}{@{}l|}{\textit{OctoThinker-8B-Hybrid-Base}} &&&&&& \\
Base Model 
& 27.5 & 22.1 & 44.2 & 68.6 & 16.7 & 35.8
& 14.7 & 11.4 & 15.7 & 0.6 & 10.6
& 24.6 \\
\quad + R-Zero 
& \underline{32.5} & 33.1 & \underline{58.4} & 85.2 & 22.6 & 46.4
& 37.4 & 17.9 & 21.7 & \underline{7.8} & 21.2
& 35.2 \\
\quad + Absolute Zero 
& \underline{32.5} & 34.9 & 56.8 & 87.0 & \underline{25.6} & 47.4
& 31.4 & 18.8 & 27.8 & 5.0 & 20.8
& 35.5 \\

\quad + SPICE
& \textbf{35.2} & \underline{40.8} & \underline{58.4} & \underline{87.3} & \underline{25.6} & \underline{49.5}
& \underline{41.3} & \underline{19.9} & \underline{29.8} & 7.2 & \underline{24.5}
& \underline{38.4} \\

\quad + {DARC (Ours)}
& 31.9 & \textbf{43.0} & \textbf{62.4} & \textbf{88.0} & \textbf{30.7} & \textbf{51.2}
& \textbf{43.8} & \textbf{22.3} & \textbf{32.3} & \textbf{10.8} & \textbf{27.3}
& \textbf{40.6} \\

\bottomrule
\end{tabular}%
}
\caption{Performance comparison on reasoning benchmarks (\%). \textbf{Bold} and \underline{underlined} values indicate the best and second-best performance within each model scale. Shaded rows indicate that \protect\colorbox{gray!10}{{General-Reasoner}} is a supervised reference and is not directly comparable to self-evolving baselines due to its different supervision regime.}
\label{main_table}
\end{table*}

\section{Experiments}

\subsection{Experimental Setup}
\paragraph{Models} 
To evaluate the generality of \method{} across model scales and architectures, we adopt representative backbones from both the Qwen and LLaMA families, including Qwen3-4B/8B-Base and OctoThinker-8B-Hybrid-Base.

\paragraph{Baselines}
We compare \method{} with representative baselines: 
(1) \textbf{Base Model}, the pretrained checkpoint without post-training; 
(2) \textbf{Absolute Zero}~\cite{zhao2025absolute}, a domain-specific grounded self-play method; 
(3) \textbf{R-Zero}~\cite{huang2025r}, a label-free self-evolving approach; 
(4) \textbf{R-Few}~\cite{yu2025guided}, a weakly supervised self-evolution method; and 
(5) \textbf{SPICE}~\cite{liu2025spice}, a corpus-grounded self-play framework.
In addition, we report \textbf{General-Reasoner}~\cite{ma2025general}, trained on approximately 230K human-annotated data, as a supervised reference for calibrating model capabilities.


We reproduce SPICE using the same prompts and corpus as \method{}, while results for other baselines are taken from \citet{yu2025guided}. For R-Few, we report its 1\% setting, which uses approximately 2.3K human-labeled WebInstruct samples.

\paragraph{Benchmarks}
We evaluate \method{} on both mathematical and general reasoning benchmarks. 
For \textbf{mathematical reasoning}, we use MATH-500~\cite{hendrycks2measuring}, GSM8K~\cite{cobbe2021training}, OlympiadBench~\cite{he2024olympiadbench}, Minerva Math~\cite{lewkowycz2022solving}, and AMC~\cite{MAA_AMC}, following the \textit{simple-evals} protocol with GPT-4o as an automatic judge. 
For \textbf{general reasoning}, we evaluate on MMLU-Pro~\cite{wang2024mmlu}, SuperGPQA~\cite{du2025supergpqa}, GPQA-Diamond~\cite{rein2024gpqa}, and BBEH~\cite{kazemi2025big}, using greedy decoding and exact-match evaluation.
Refer to Appendix \ref{sec::exp_detail} and \ref{sec::prompt_design} for experimental details.
\begin{figure*}[t]
    \centering
    \begin{subfigure}[t]{0.32\textwidth}
        \centering
        \includegraphics[width=\linewidth]{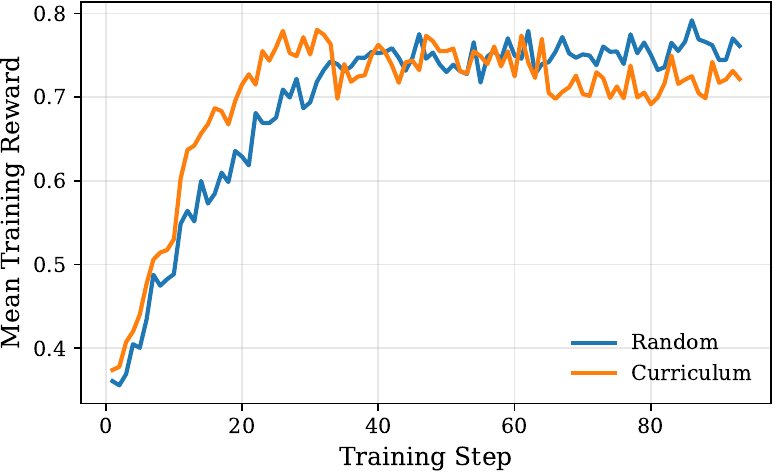}
        \caption{Training reward.}
        \label{fig:train_reward_curve}
    \end{subfigure}
    \hfill
    \begin{subfigure}[t]{0.32\textwidth}
        \centering
        \includegraphics[width=\linewidth]{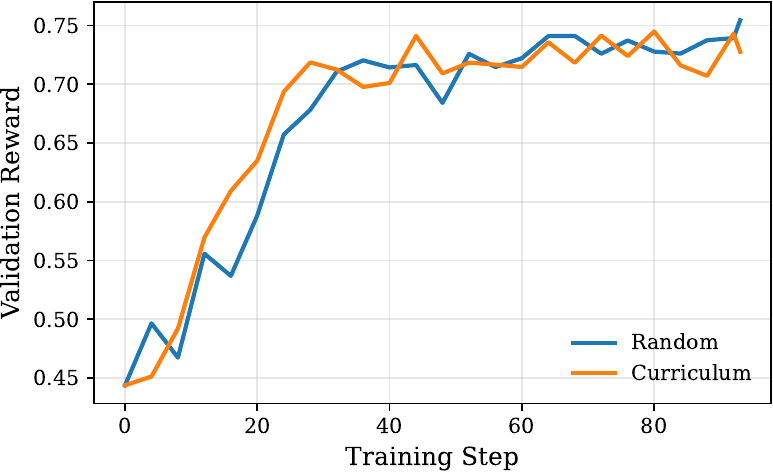}
        \caption{Validation reward.}
        \label{fig:val_reward_curve}
    \end{subfigure}
    \hfill
    \begin{subfigure}[t]{0.32\textwidth}
        \centering
        \includegraphics[width=\linewidth]{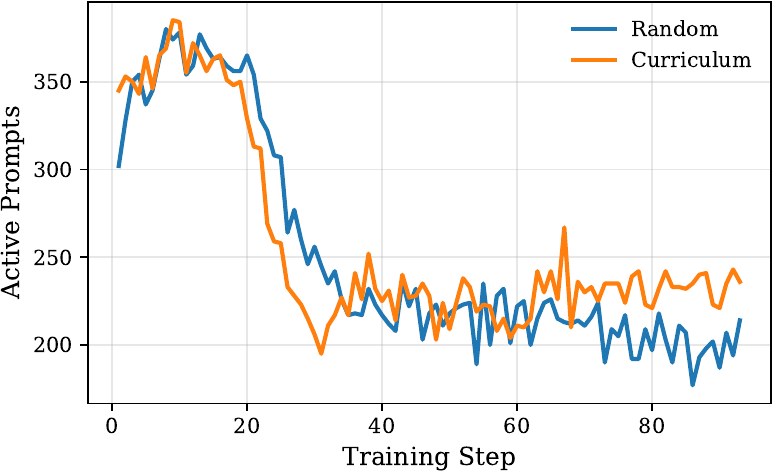}
        \caption{Active prompts in GRPO.}
        \label{fig:active_prompts_curve}
    \end{subfigure}

\caption{Training dynamics of \method{} under curriculum-ordered training and random shuffling on the same offline question set. The validation reward in (b) is evaluated on the Math12K test set.} 

    \label{fig:curriculum_vs_shuffle_dynamics}
\end{figure*}

\subsection{Main Results}

Table~\ref{main_table} summarizes the results on both mathematical and general-domain reasoning benchmarks. We draw three observations.

\noindent\textbf{First, self-evolution consistently improves reasoning, with larger gains on weaker backbones and stronger gains on math.}
Across all backbones, self-evolving methods improve over the corresponding base checkpoints, with larger relative gains on initially weaker models such as OctoThinker. Improvements are also typically larger on mathematical reasoning than on general reasoning, likely due to more deterministic supervision and clearer correctness signals in math evaluations.

\noindent\textbf{Second, \method{} is the strongest among label-free self-evolving baselines and remains competitive with weakly supervised self-play.}
\method{} consistently outperforms all label-free baselines including R-Zero, Absolute Zero, and SPICE on both average score of mathematical and general reasoning benchmarks across all three backbones. Relative to weakly supervised approaches such as R-Few, \method{} remains competitive while requiring no human annotations.
On average, \method{} improves the average score by \textbf{10.9 points} over the base models, validating the effectiveness of our decoupled framework without tightly coupled co-evolution.

\noindent\textbf{Third, \method{} can approach supervised pipelines at sufficient model scale.}
With Qwen3-8B as the backbone model, label-free \method{} matches the overall average performance of the supervised General-Reasoner. This indicates that strong base models can achieve competitive reasoning performance through self-evolution alone, without relying on annotation-intensive supervised training.

\subsection{Analysis \& Discussion}
\paragraph{\Solver{} Training dynamics.}
We analyze the \Solver{} training dynamics to assess training stability.
As shown in Figure~\ref{fig:curriculum_vs_shuffle_dynamics}(a), the training reward increases sharply at the beginning and then gradually saturates.
We observe two transient drops around steps 32 and 64, which coincide with scheduled curriculum transitions from Easy to Medium, and from Medium to Hard, respectively.
This interpretation is corroborated by Figure~\ref{fig:curriculum_vs_shuffle_dynamics}(c), where the number of active prompts rises at the same steps, suggesting that the optimization is exposed to a more challenging prompt set.
Importantly, as shown in Figure~\ref{fig:curriculum_vs_shuffle_dynamics}(b), the validation reward exhibits a steady upward trend without degradation, contrasting with the training collapse reported in prior self-play systems.
Overall, these dynamics provide evidence that \method{} improves the stability of LLM evolution by maintaining a more reliable training signals. Analysis of \Questioner{} training dynamics is listed in Appendix \ref{app:questioner_dynamics}.

\paragraph{Effect of Asymmetric Self-Distillation.}
To assess the efficacy of asymmetric distillation, we compare document-grounded prompting (questions + documents) against question-only prompting, measuring relative gains via Avg@8 win rates.
Table~\ref{tab:text_length_analysis} shows that document augmentation consistently yields win rates $>50\%$ in short- and medium-context regimes, confirming its utility for supervision. Notably, Qwen2.5-7B-Instruct outperforms the base model, indicating that instruction tuning improves evidence extraction and noise resilience. 
However, gains diminish in long-context scenarios ($>5$K tokens), likely due to  excessive document length introduces extraneous noise that may dilute the supervisory signal. 
These results validate our asymmetric distillation framework while highlighting document length as a key constraint.



\begin{table}[h]
\centering
\small
\setlength{\tabcolsep}{6pt}
\begin{tabularx}{\columnwidth}{lCCCC}
\toprule
\textbf{Model} & \textbf{Short} & \textbf{Med.} & \textbf{Long} & \textbf{Avg.} \\
\midrule
Qwen3-4B-Base & 52.9 & 54.8 & 39.2 & 50.9 \\
Qwen2.5-7B-Instruct & 54.5 & 60.0 & 48.7 & 53.9 \\
\bottomrule
\end{tabularx}
\caption{Avg@8 win rate (\%) of document-augmented prompting over non-augmented prompting on discrepant instances, grouped by document lengths. }
\label{tab:text_length_analysis}
\end{table}

\paragraph{Cross-model consistency of question difficulty.} \label{sec::cross_model_consistency}

\begin{figure}[t]
    \centering
    \includegraphics[width=\columnwidth]{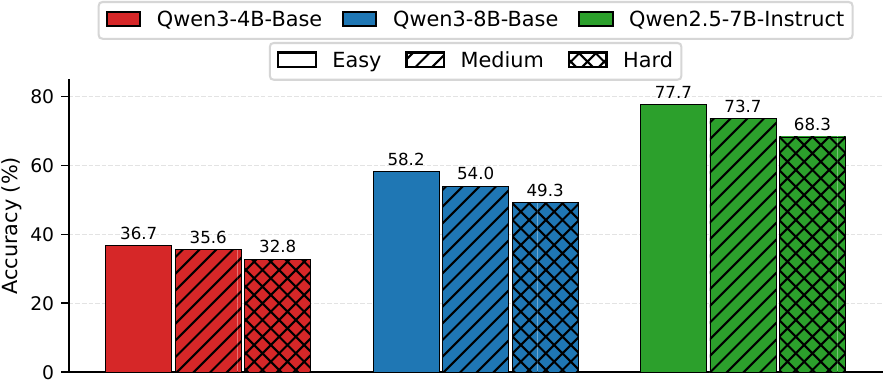}
    \caption{Accuracy of different \Solver{}s on questions generated under different difficulty levels.}
    \label{fig::accuracy_by_difficulty}
\end{figure}

To evaluate whether reinforcement learning enables the \Questioner{} to generate questions of different difficulty, we measure various \Solver{}s' average accuracy.
As shown in Figure~\ref{fig::accuracy_by_difficulty}, all \Solver{} accuracy decreases monotonically from Easy to Hard across all models, indicating that different input conditions naturally give rise to progressively harder questions.
Notably, this monotonic trend is preserved across backbones, suggesting that the induced difficulty ordering is largely independent of the \Solver{} backbone used.
These results indicate that the \Questioner{} learns a solver-agnostic and input-driven difficulty partition. 
Appendix~\ref{sec:case_study_of_questions} presents a case study of the generated questions.

\paragraph{Cross-solver generalization of the \Questioner{}.}
To evaluate whether the learned \Questioner{} generalizes beyond the \Solver{} backbone used during training, we reuse the question set generated by the Qwen3-4B-Base \Questioner{} to train different \Solver{}s.
As shown in Table~\ref{tab:cross-solver}, both a larger 8B model and a smaller 1.7B model achieve consistent performance improvements, demonstrating robust cross-solver generalization.
Moreover, further tuning the trained models on human-annotated data using our synthesized question set yields additional performance gains, indicating that the self-evolving curriculum is complementary to human supervision.
These results validate our core design: decoupled question generation creates versatile, reusable curriculums that benefit heterogeneous \Solver{}s without overfitting to a specific backbone.
\begin{table}[t]
\centering
\small
\setlength{\tabcolsep}{3pt}  
\begin{tabularx}{\columnwidth}{@{}XYY@{}}
\toprule
\textbf{Model} & \textbf{Math Avg.} & \textbf{General Avg.} \\
\midrule
\textit{Qwen3-1.7B-Base} & 49.6 & 21.3 \\
\quad + DARC
  & 51.4 {\scriptsize\textcolor{gray}{(+1.8)}}
  & 25.9 {\scriptsize\textcolor{gray}{(+4.6)}} \\
\midrule
\textit{Qwen3-8B-Base} & 63.3 & 33.1 \\
\quad + DARC
  & 70.9 {\scriptsize\textcolor{gray}{(+7.6)}}
  & 37.5 {\scriptsize\textcolor{gray}{(+4.4)}} \\
\midrule
\textit{Qwen3-4B-Base (Human)} & 65.5 & 33.2 \\
\quad + DARC
  & 66.7 {\scriptsize\textcolor{gray}{(+1.2)}}
  & 34.4 {\scriptsize\textcolor{gray}{(+1.2)}} \\
  
\bottomrule
\end{tabularx}
\caption{Experimental results of \method{} with backbones different from \Questioner{} training (\%).
Qwen3-4B-Base (Human) indicates GRPO tuning on the human-annotated Math12K training dataset.}
\label{tab:cross-solver}
\end{table}

\paragraph{\method{} Improves reasoning beyond corpus memorization.} To examine whether \method{} enhances reasoning capabilities beyond merely memorizing a task-specific corpus, we compare it against a baseline fine-tuned on the same data using the standard next-token prediction objective (Vanilla FT). As shown in Table~\ref{tab:darc_beyond_memorization}, while both methods improve upon the base models, \method{} consistently achieves larger gains across model scales and domains. Notably, although both methods exhibit diminishing absolute gains as model size increases, Vanilla FT saturates faster. Consequently, the performance gap between \method{} and Vanilla FT widens from an average of \textbf{3.3} points on the 4B model to \textbf{3.9} points on the 8B model. These results indicate that \method{} more effectively exploits the corpus to induce transferable reasoning abilities rather than surface-level pattern memorization.

\begin{table}[t]
\centering
\small
\setlength{\tabcolsep}{3pt}  
\begin{tabularx}{\columnwidth}{@{}XYY@{}}
\toprule
\textbf{Model} & \textbf{Math Avg.} & \textbf{General Avg.} \\
\midrule
\textit{Qwen3-4B-Base} & 53.1 & 27.9 \\
\quad + Vanilla FT
  & 62.9 {\scriptsize\textcolor{gray}{(+9.8)}}
  & 31.3 {\scriptsize\textcolor{gray}{(+3.4)}} \\
\quad + DARC
  & 66.7 {\scriptsize\textcolor{gray}{(+13.6)}}
  & 34.1 {\scriptsize\textcolor{gray}{(+6.2)}} \\
\midrule
\textit{Qwen3-8B-Base} & 63.3 & 33.1 \\
\quad + Vanilla FT
  & 65.5 {\scriptsize\textcolor{gray}{(+2.2)}}
  & 35.5 {\scriptsize\textcolor{gray}{(+2.4)}} \\
\quad + DARC
  & 71.1 {\scriptsize\textcolor{gray}{(+7.8)}}
  & 37.8 {\scriptsize\textcolor{gray}{(+4.7)}} \\
\bottomrule
\end{tabularx}
\caption{Comparison between \method{} and finetuning with next-token prediction on the same corpus (\%).}
\label{tab:darc_beyond_memorization}
\end{table}

\begin{table}[htbp]
\centering
\small
\setlength{\tabcolsep}{4pt}
\begin{tabularx}{\columnwidth}{@{}Xcc@{}}
\toprule
\textbf{Method} & \textbf{Math Avg.} & \textbf{General Avg.} \\
\midrule
\method{} & 66.7 & 34.1 \\
\midrule
\textit{Ablations} & & \\
\quad w/o Asymmetric Distillation
& 65.0 {\scriptsize\textcolor{gray}{(-1.7)}} & 33.6 {\scriptsize\textcolor{gray}{(-0.5)}} \\
\quad w/o Specialized \Questioner{}
& 65.5 {\scriptsize\textcolor{gray}{(-1.2)}} & 32.7 {\scriptsize\textcolor{gray}{(-1.4)}} \\
\quad w/o Difficulty Awareness
& 65.3 {\scriptsize\textcolor{gray}{(-1.4)}} & 32.7 {\scriptsize\textcolor{gray}{(-1.4)}} \\
\bottomrule
\end{tabularx}
\caption{Ablation study (\%).
\emph{w/o Asymmetric Distillation}: majority-voting pseudo-labels.
\emph{w/o Specialized \Questioner{}}: generic model Qwen3-4B as \Questioner{}.
\emph{w/o Difficulty Awareness}: easy questions only training.}
\label{tab:rzero_ablation}
\end{table}

\paragraph{Ablation Study.}
We conduct an ablation study to isolate the contribution of each core component in \method{}.
As reported in Table~\ref{tab:rzero_ablation}, removing any of the components consistently degrades performance, indicating that these modules are complementary.
Notably, replacing our trained \Questioner{} with a generic strong model Qwen3-4B also yields worse results, suggesting that the learned, difficulty-aware \Questioner{} provides a more effective curriculum.

\section{Conclusion}

In this work, we introduce the \method{} framework, which adopts decoupled training and asymmetric self-distillation to stabilize self-evolving. Extensive results suggest that \method{} consistently outperforms existing baselines. We hope this work provides useful insights for LLM self-evolution.

\section*{Limitations}
While the proposed framework  represents a notable step toward stable LLM self-evolution, it still has several limitations.
First, \method{} relies on an external corpus to ground both the \Questioner{} and the \Solver{}, which constrains its applicability in fully data-free scenarios.
Second, the pseudo-labels produced via asymmetric self-distillation are inevitably noisy, potentially limiting further performance gains.
Third, the current framework is primarily designed for domains with verifiable answers, which restricts its applicability to open-ended tasks.
We leave addressing these limitations to future work.

\section*{Ethical Statement}
This work investigates self-evolution of large language models purely as a research methodology, without deployment in real-world or user-facing scenarios. The models operate in an offline experimental environment and are not used for autonomous decision-making. Therefore, potential risks are minimal and largely limited to methodological concerns.

We use only publicly available corpora (e.g., DataComp-LM and Nemotron-CC-Math) and do not collect any user data. We rely on the datasets’ documented curation/filtering procedures to reduce personally identifying information (PII) and offensive content. We do not release raw training text or analyze any information at the level of individual people; we only report aggregated benchmark results. Any remaining PII/offensive content in web-scale corpora is treated as an inherent limitation.

ChatGPT was used solely to assist with language refinement and improving the clarity of presentation. It did not contribute to the development of ideas, experimental design, data analysis, interpretation of results, or drawing scientific conclusions.

\bibliography{custom}

\appendix

\section{Experiment Details}\label{sec::exp_detail}

This section presents the detailed hyperparameters and configurations used in our experiments.

\subsection{Common Experimental Setup}

Our experiments are conducted on 8 NVIDIA A800 (80GB) GPUs, using the \texttt{veRL}~\cite{sheng2025hybridflow} framework. 
We employ \texttt{vLLM}~\cite{kwon2023efficient} to facilitate efficient inference, utilizing a tensor parallel size of 2 to handle large-scale rollouts. The models are optimized via AdamW~\cite{loshchilov2019adamw} with a learning rate of $1\text{e-}6$ and weight decay of $1\text{e-}2$. We choose Group Relative Policy Optimization (GRPO) as our reinforcement learning algorithm.  To accommodate extensive corpus and problem descriptions, the maximum prompt length is set to 8,192 tokens and the generation limit is set to 4,096 tokens.

\subsection{Questioner Training Details}
During \Questioner{} training phase, we source 10,000 documents from 
Nemotron-CC-Math~\cite{mahabadi2025nemotron} and 10,000 documents from 
DataComp-LM~\cite{li2024datacomp}, respectively.
Training is conducted for 1 epoch with a global batch size of 16. For each document-difficulty pair, the model samples $G=8$ candidate questions.  And for each question, the privileged \Solver{} samples $N=8$ trajectories to get the pseudo-label.
For each backbone reported in Table~\ref{main_table}, we use its corresponding base model as the \Solver{} when computing the empirical success rate (Eq.~\ref{empirical_success_rate}), ensuring a fair comparison with prior work. 
We further show that the relative difficulty of questions exhibits strong ranking consistency across different \Solver{}s (Section~\ref{sec::cross_model_consistency}). 
Difficulty is calibrated by setting the target parameter $\tau$ to $0.8$, $0.5$, and $0.2$, corresponding to easy (80\% accuracy), medium (50\%), and hard (20\%) difficulty tiers, respectively.

In the \Questioner{} training stage, the judge model is used only to provide binary feedback on whether a generated question is grounded in the given document, which does not constitute answer labels or knowledge distillation. Our experiments show that any model with basic instruction-following ability can serve as the judge. We initially attempted to use pretrained base models as judges; however, without instruction tuning, they often violate the required output format and produce invalid structured responses. Therefore, without loss of generality, we adopt Qwen2.5-7B-Instruct as the judge in all reported experiments. This choice is motivated by implementation stability and cost-effectiveness rather than a methodological dependence on a stronger external teacher. Alternative implementations, such as ROUGE-based heuristics or embedding-similarity matching, are equally compatible with \method{}. Accordingly, the LLM-as-a-Judge component offers limited guidance and is not essential to the label-free nature of the proposed self-evolution framework.

\subsection{Solver Training Details}
The \Solver{} is trained on questions generated by the \Questioner{}. 
We generate up to $60{,}000$ candidate questions (20{,}000 per difficulty level), and retain only those that satisfy the required output format. 
The resulting set of valid questions defines the final training set of size $M$, without any additional filtering.
To balance training efficiency and pseudo-label reliability, we set the number of rollouts $N=8$ and the acceptance threshold $\gamma=0.3$. 
The privileged teacher \Solver{} shares parameters with the target student \Solver{}, i.e., the teacher is continuously updated alongside training and reflects the current state of the student model.
Optimization is performed with a global batch size of 512 for a single training epoch. 
During training-time inference, we use a temperature of 1.0 and top-$p$ sampling with $p=0.99$.

\section{Effect of Curriculum Learning in Training \Solver{} }
We isolate the impact of curriculum ordering by comparing it against a random shuffling strategy on the same offline question set. As shown in Figure~\ref{fig:curriculum_vs_shuffle_dynamics}(a), curriculum learning significantly enhances early-stage sample efficiency. Specifically, the model reaches a validation reward threshold of $0.7$ in just 24 steps, compared to 32 steps for random shuffling. Moreover, we found that this efficiency gain is not due to simply accessing more active prompts. The number of active prompts remains comparable between the two settings, as shown in Figure~\ref{fig:curriculum_vs_shuffle_dynamics}(c). 
This suggests that curriculum ordering improves early-stage alignment between the \Solver{} and question difficulty, enabling smoother optimization and faster initial progress while preserving robust final performance.

\section{Training Dynamics of the \Questioner{}}
\label{app:questioner_dynamics}

\begin{figure}[h]
    \centering
    \begin{subfigure}[h]{\columnwidth}
        \centering
        \includegraphics[width=0.9\linewidth]{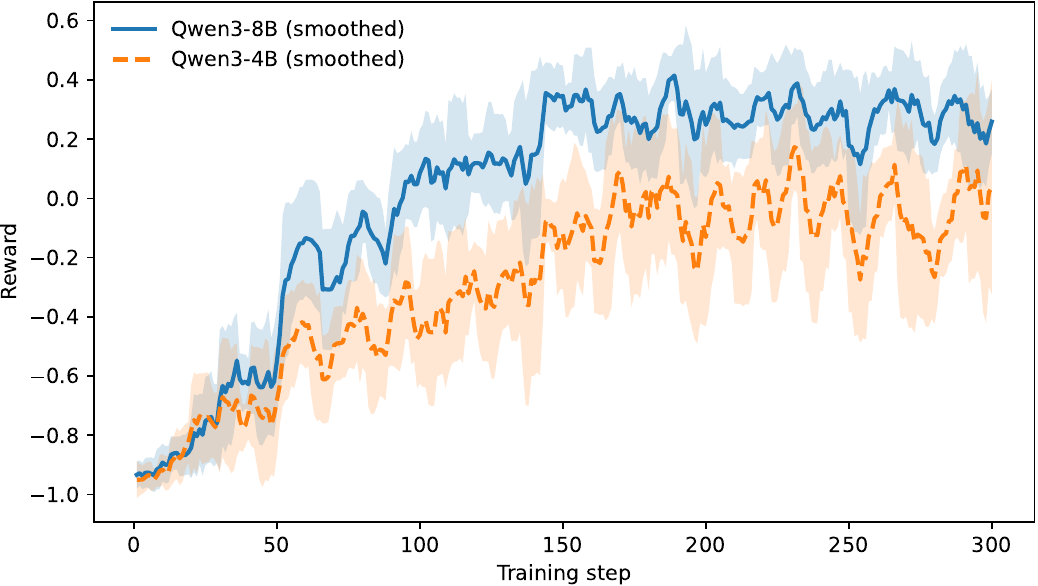}
        \caption{\Questioner{} reward.}
        \label{fig:questioner_reward_dyn}
    \end{subfigure}

    \vspace{0.6em}

    \begin{subfigure}[t]{\columnwidth}
        \centering
        \includegraphics[width=0.9\linewidth]{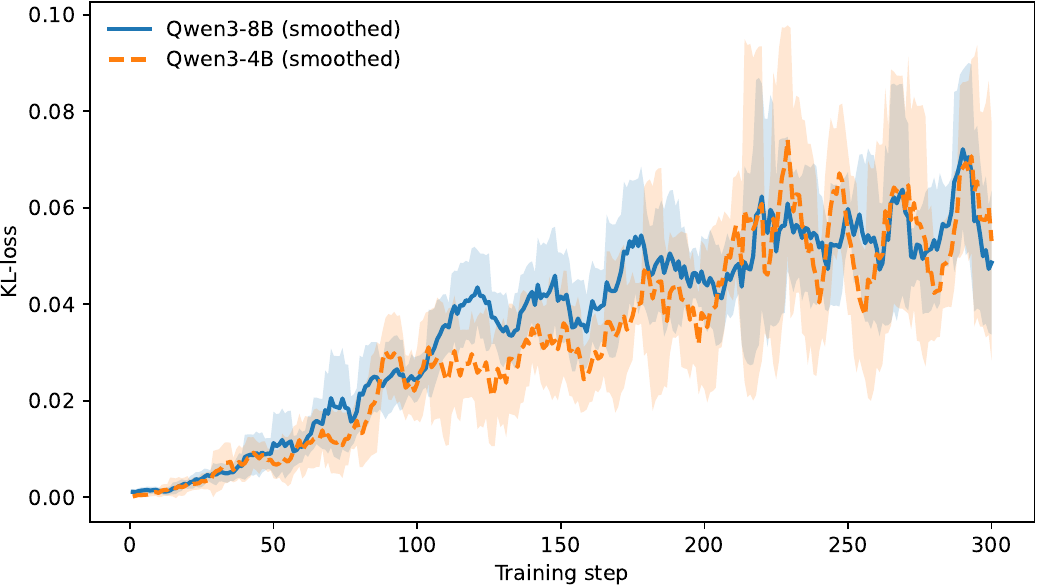}
        \caption{KL loss during GRPO.}
        \label{fig:questioner_kl_dyn}
    \end{subfigure}

    \caption{
    Training dynamics of the \Questioner{} for Qwen3-4B and Qwen3-8B backbones.
    Curves are smoothed for readability; shaded regions indicate within-run variability measured by a rolling standard deviation over steps.
    }
    \label{fig:questioner_dynamics}
\end{figure}

We examine the training dynamics of the \Questioner{} as an additional diagnostic of optimization behavior.
Figure~\ref{fig:questioner_dynamics} reports (a) the \Questioner{} reward and (b) the KL loss during GRPO training.
In both model scales, the reward increases rapidly in the early stage and then plateaus, while the KL loss rises gradually before stabilizing.
Overall, these trends suggest a stable training process, with the optimization dynamics gradually approaching a steady regime.
We also observe that the 8B \Questioner{} attains a higher reward and converges faster than the 4B \Questioner{}, suggesting that larger models can more readily adapt their generation to match the specified difficulty level.

\section{Case Study of Generated Questions}
To qualitatively illustrate how the difficulty level affects question generation, we sample a single input document and prompt the trained \Questioner{} with three difficulty conditions (Easy/Medium/Hard).
Figure~\ref{fig:example_questions} shows that the \Questioner{} produces questions with progressively higher solving difficulty and self-contained questions without access to the source document.
Specifically, the Easy question is more direct and relies on shallow reasoning with clearer dis-tractors, whereas the Medium and Hard questions introduce more nuanced constraints and more competitive dis-tractors, requiring the \Solver{} to disambiguate closely related alternatives based solely on the problem statement.

\label{sec:case_study_of_questions}

\noindent

\begin{figure*}[t] 
\centering
\begin{redexampleFloatingTitle}
{Questions at Different Difficulty Levels}
\textbf{Document:} 

``Monitoring flows at routers for flow analysis or deep packet inspection requires the monitors to update monitored flow information at the transmission line rate and needs to use high-speed memory such as SRAM. Therefore, it is difficult to measure all flows, and the monitors need to limit the monitoring target to a part of the flows. However, if monitoring targets are randomly selected, an identical flow will be monitored at multiple routers on its route, or a flow will not be monitored at any routers on its route. To maximize the number of flows monitored in the entire network, the monitors are required to select the monitoring targets while maintaining a balanced load among the monitors...''

\medskip
\textbf{Generated Questions:}  

\textbf{Easy}
\\The primary goal in flow monitoring at routers is to: \\A. Maximize the use of high-speed memory. \\B. Ensure every flow is monitored. \\C. Limit the monitoring to prevent redundant or missed flows while balancing the load. \\D. Monitor all flows but avoid using high-speed memory.\\
Answer: \textbf{C}

\medskip

\textbf{Medium}
\\What is the primary reason for ensuring a balanced load among routers in the context of flow monitoring for deep packet inspection?\\A. To prevent identical flows from being monitored at multiple routers.\\B. To ensure that no flows are overlooked by routers.\\C. To reduce the load on routers with high-speed memory.\\D. To maximize the efficiency of flow measurement across the entire network.\\
Answer: \textbf{D}
\medskip

\textbf{Hard}
\\Given the context that routers use SRAM for high-speed memory to monitor flows at transmission line rate for flow analysis or deep packet inspection, which option best describes the challenge and solution for effective flow monitoring?\\A. SRAM is used to load balance among monitors without considering flow target.\\B. Flow monitoring in SRAM is limited by the need to randomly select monitoring targets.\\C. Load balancing and efficient target selection are crucial for maximizing global flow monitoring in SRAM.\\D. Flow monitoring in SRAM requires frequent memory updates to target every flow equally.\\E. Monitoring targets should be limited to one flow per router to avoid SRAM overload.        \\
Answer: \textbf{C}
\end{redexampleFloatingTitle}
  \captionof{figure}{Case Study of generated questions with different difficulty levels.}
  \label{fig:example_questions}
\end{figure*}

\section{Prompt Design} \label{sec::prompt_design}
We present the prompt used for the \Solver{} and \Questioner{} in Figure \ref{fig:sovler_prompt} and Figure \ref{fig:questioner_prompt}, respectively.


\begin{figure*}[t]
\centering
    \begin{redexampleFloatingTitle}
    {Solver Prompt Template}
    \textbf{For Student Solver:}
    \\Please reason step by step, and put your final answer option within \texttt{//}boxed\texttt{\{\}}. Only put the letter in the box, e.g. \texttt{//}boxed\texttt{\{A\}}. There is only one correct answer.
\texttt{\{Question\}}
    
    \medskip
        
    \textbf{For Teacher Solver:}
\\Read the following context and answer the question. \texttt{\{Document\}} Please reason step by step, and put your final answer option within \texttt{//}boxed\texttt{\{\}}. Only put the letter in the box, e.g. \texttt{//}boxed\texttt{\{A\}}. There is only one correct answer.
\texttt{\{Question\}}
    
    \medskip
    
\textit{Note: \texttt{\{Question\}} and \texttt{\{Document\}} are placeholders for the actual problem and document, respectively.}
    \end{redexampleFloatingTitle}
    
    \captionof{figure}{The prompt used for the \Solver{} model.} 
    \label{fig:sovler_prompt}
\end{figure*}

\begin{figure*}[t]
\centering
    \begin{redexampleFloatingTitle}
    {Questioner Prompt Template}
\setlength{\parindent}{0pt}
Your task is to generate a single self-contained question and its correct answer inspired by the given document.
The question must strictly satisfy both the difficulty level and the answer\_type constraints.
You must output exactly one JSON object as specified below.
All reasoning MUST be placed inside the ``analysis'' field of the JSON.

\medskip
Difficulty Level\\
You are given a target difficulty level: \texttt{\{Difficulty ID\}}

You must follow these operational definitions: \texttt{\{Difficulty ID Definitions\}}
\\Your generated question and solution process must match the target difficulty level as closely as possible.

\medskip
Answer Type\\
You must generate a question whose answer has the following type:
\texttt{\{Answer Type\}}

Rules:
\texttt{\{Answer Type Definitions\}}

\medskip
Core Requirements for the Question
\begin{itemize}\itemsep0pt \parskip0pt
  \item The question must be inspired by the document (but self-contained).
  \item The question must not reference ``the document'' or ``the text''.
  \item The question must be understandable by someone who only sees the question.
  \item The reasoning steps must match the difficulty level.
  \item The question must combine the number of spans required by difficulty level.
  \item The answer must be unique and consistent with the document.
  \item All variables must be defined in the question itself.
  \item No ambiguity.
  \item Don't copy the question from the document.
\end{itemize}

\medskip
Final Output Format (STRICT)\\
Your output must be exactly one JSON object with the following fields:
\begin{itemize}\itemsep0pt \parskip0pt
  \item ``analysis'' (string)
  \item ``question'' (string)
  \item ``intermediate\_results'' (object)
  \item ``answer'' (string)
  \item ``solving\_time\_estimate'' (number)
  \item ``required\_concepts'' (array of strings)
  \item ``potential\_errors'' (array of strings)
\end{itemize}

\medskip
Field Specifications
\begin{itemize}\itemsep0pt \parskip0pt
  \item ``analysis'' (string): full internal reasoning (spans, difficulty mapping, design, calculations, uniqueness, answer\_type).
  \item ``question'' (string): one self-contained exam-style question; no reasoning/hints/metadata.
  \item ``intermediate\_results'' (object): short step names $\rightarrow$ 1--20 sentence summaries.
  \item ``answer'' (categorical): one uppercase letter.
  \item ``solving\_time\_estimate'' (number): minutes.
  \item ``required\_concepts'' (array of strings): 1--10 items.
  \item ``potential\_errors'' (array of strings): 1--10 items.
\end{itemize}

\medskip
Example (do not copy; only follow structure)\\
The example below is for structure demonstration only. Your output must be based on the provided \texttt{\{Document\}} above and strictly follow the assigned difficulty\_id and answer\_type. Example Input: \texttt{\{Example Input\}}

\medskip
\textit{Note: \texttt{\{Document\}}, \texttt{\{Example Input\}}, \texttt{\{Difficulty ID\}} ,\texttt{\{Difficulty ID Definitions\}} , \texttt{\{Answer Type\}} and \texttt{\{Answer Type Definitions\}}) are placeholders for their corresponding actual content.}

    \end{redexampleFloatingTitle}
    
    \captionof{figure}{The prompt used for the \Questioner{} model.} 
    \label{fig:questioner_prompt}
\end{figure*}

\end{document}